\pdfoutput=1
\documentclass[aps, prb, 12pt,preprint,tightenlines,nofootinbib]{revtex4}
\usepackage[toc,page]{appendix}
\usepackage{appendix}
\usepackage{epsfig}
\usepackage{hyperref}
\usepackage{amssymb}
\usepackage{color}
\usepackage{wrapfig}
\usepackage{caption}
\usepackage{slashed}
\usepackage{amsmath}
\usepackage{mathptmx}
\newcommand{\be}{\begin{equation}}
\newcommand{\ee}{\end{equation}}
\newcommand{\ba}{\begin{eqnarray}}
\newcommand{\ea}{\end{eqnarray}}
\newcommand{\bd}{\begin{displaymath}}
\newcommand{\ed}{\end{displaymath}}

\newcommand{\commentout}[1]{{}}

\begin{document}

\bibliographystyle{naturemag} 

\title{
Determining Health Utilities through \\ Data Mining of Social Media\vspace{.2cm}}
\author{CL Thompson}
\author{Josh Introne}
\author{Clint Young}

\vspace{-80mm}

\date{\today}

\begin{abstract}

`Health utilities' measure patient preferences for perfect health compared to specific unhealthy states, such as asthma, a fractured hip, or colon cancer. When integrated over time, these estimations are called quality adjusted life years (QALYs). Until now, characterizing health utilities (HUs) required detailed patient interviews or written surveys. While reliable and specific, this data remained costly due to efforts to locate, enlist and coordinate participants. Thus the scope, context and temporality of diseases examined has remained limited.
					
Now that more than a billion people use social media, we propose a novel strategy: use natural language processing to analyze public online conversations for signals of the \textit{severity} of medical conditions and correlate these to known HUs using machine learning. In this work, we filter a dataset that originally contained 2 billion tweets for relevant content on 60 diseases. Using this data, our algorithm successfully distinguished mild from severe diseases, which had previously been categorized only by traditional techniques. This represents progress towards two related applications: first, predicting HUs where such information is nonexistent; and second, (where rich HU data already exists) estimating temporal or geographic patterns of disease severity through data mining. 
	
\end{abstract}

\maketitle

\section{Full Text:}	
	The game theorist John Von Neumann and his collaborator, Morgenstern, designed one of the earliest measures of health utility (HU).\cite{neumann_theory_1944}  Their method, called the Standard Gamble, quantifies quality-of-life by first asking patients to make a hypothetical, important decision. Ultimately, the information contained in that decision is converted to a number ranging between 0 and 1. To start, a patient imagines that researchers have developed a potent drug that can cure the patient's disease. However the pill has a terrible side effect in some patients-- instant death. So the patient is asked to decide on the maximum acceptable risk (m), expressed as a probability, that would allow proceeding with the therapy. A HU (u) is defined as 1 - m = u.   
	
\begin{wrapfigure}{r}{0.45\textwidth}
\vspace{-5mm}
\captionsetup{font=scriptsize}
\includegraphics[width = .45\textwidth]{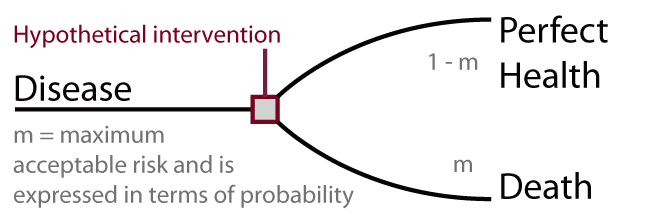}	
\caption{Standard Gamble}	
\vspace{-4.8mm}
\end{wrapfigure}
	
Patients with a lower quality of life (lower u) have a stronger desire for health and accommodate more risk (higher m). Importantly, health utilities are not disease-specific, so diverse health conditions and outcomes can be compared. For example, we can say that epilepsy is worse than asthma because, on average, epilepsy patients accept a higher risk of death when faced with the dilemma in the Standard Gamble. In contrast, there is no clear hierarchy for \textit{disease-specific} endpoints such as seizure frequency or pulmonary function.

The integral of a HU, with respect to time, is called a Quality Adjusted Life Year (QALY). This is a metric widely used in cost-utility analyses which assess relative economic value of various health interventions.  Therefore, thousands of published studies (including clinical guidelines and CDC assessments of public health) rely on estimates of disease severity in the form of HUs.\cite{neumann_legislating_2010}

For decades, questionnaires or interviews like the Standard Gamble have served to elicit health utilities. Other examples in this class include: Time Trade Off and the Health Utilities Index. An alternative involves using algorithms that map clinical data to HUs. Generally, for a given health condition, clinical data and HU data must be collected simultaneously; then a regression model is built such that the objective data predicts the HU. For example, epilepsy patients with more frequent seizures may report lower quality-of-life, (although some evidence confounds this assumption).\cite{choi_seizure_2014} These mappings are generally limited to their disease context. For example, knowledge about the relationship of seizure frequency to HU does not transfer to a mapping of pulmonary function to HU.

Most recently, investigators have utilized natural language processing to filter Twitter and other social networks for novel indices of disease severity. Generally, HUs have not been utilized as training labels for these metrics, and algorithms apply to specific conditions only, such as toothaches\cite{heaivilin_public_2011}  and depression.\cite{de_choudhury_social_2013} Despite vast repositories of patient generated data, there is no general data mining approach that can estimate a HU for any given disease.

Substantial progress has been made by Parimbelli \textit{et al.}\cite{parimbelli_use_2014} who have used sentiment analysis (SA) of online health messaging boards to derive HUs. SA uses natural language processing (NLP) to extract emotional information from online messages; corporations have used it to monitor brand appeal and word-of-mouth recommendations by consumers. Parimbelli \textit{et al.} use NLP to identify relevant content within five wellness domains (e.g. mobility, presence of pain, \textit{etc.}). Then, using SA, a negativity score is created for each domain, providing weights for its HU estimate. The application computes a final score by weighting these HUs with traditionally derived HUs. So far the system has been validated on a scenario related to atrial fibrillation.

Like Parimbelli, we have an ambitious scope: we train an algorithm to estimate a HU for potentially any disease-- not just a mapping that transforms disease-specific measurements into HUs. In contrast to Parimbelli, we use alternate natural language features to estimate HUs. We also highlight the performance of our method, comparing its estimates to held out data on 20 traditionally-derived HUs.

\begin{wrapfigure}{l}{0.27\textwidth}
\vspace{-4mm}
\captionsetup{font=scriptsize}
\includegraphics[width = .27\textwidth]{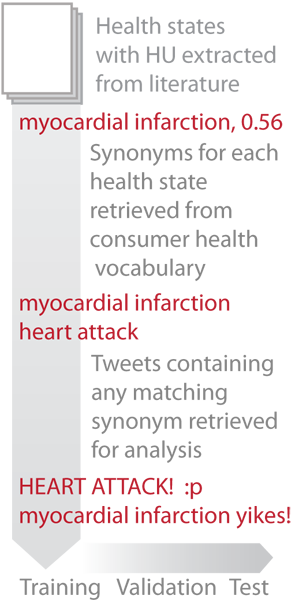}	
\caption{Process}	
\vspace{-5.7mm}
\end{wrapfigure}

To test the idea that HUs can be derived from online health conversations, we attempted to build an algorithm to classify mild versus severe health conditions-- the most essential dichotomy. We began by compiling a list of diseases with HUs previously determined by well-designed, highly-cited studies utilizing the traditional approach of interviews or surveys.

We preferred highly cited studies with the most scope in order to reduce the variation that inevitably arises between HU studies. Some studies undertake broad data collection efforts, apply a uniform methodology and report HUs for more than 100 diseases. Also, we required sufficient specificity with respect to the health conditions studied. Some HU studies aggregate conditions by International Classification of Disease code. Unfortunately, this lossy process results in terms that are too broad or vague for our purposes, such as ``esophageal problem."\cite{gold_toward_1998}

For selected studies, we extracted health conditions with their HUs as `gold-standard' labels. The conditions with the lowest HU became the `severe' set, while the upper half became the `mild' conditions, although we culled a margin of middle range of HUs (see table 2 below). Then for each condition in each class, we found related social media messages. We performed the search for related messages by using the exact term cited in the study, and then expanded our query to include all matching synonyms as provided by a consumer health vocabulary containing validated mappings between expert and layperson terminology.\cite{zeng-treitler_making_2007} We defined the entire corpora of social media messages related to a single disease, along with its class label, as `an example'.

Our social media content came from Twitter, a `microblogging' site that limits user messages (called, `tweets') to 140 characters. Challenges of using Twitter include its unmoderated content, unbounded by topic or context. Yet, users generate more than 500 million tweets every day and Twitter data has been utilized for hundreds of epidemiological studies. Data has been published showing that people share highly personal information.\cite{lee_what_2014} We obtained a dataset of 11 million health-related tweets, previously filtered from a dataset of two billion tweets.\cite{paul_discovering_2014}

Finally we followed a standard methodology for machine learning experiments: each illness was randomly assigned to training, validation, and test sets. All sets had roughly the same proportion of mild and severe illnesses.

To develop a candidate set of features, we took two complementary approaches. For the first approach, we inspected a small sample of tweets. Scrolling through the data revealed that topics related to mild health problems often included a frivolous style such as usage of emoticons, while severe diseases engendered a more formal, grave tone that included conventional punctuation and syntax. Therefore, our candidate features became emoticons and patterns related to punctuation or capitalization.
\begin{table}
\vspace{-12mm}
\begin{center}
\captionsetup{font=scriptsize}
\includegraphics[width = .96\textwidth]{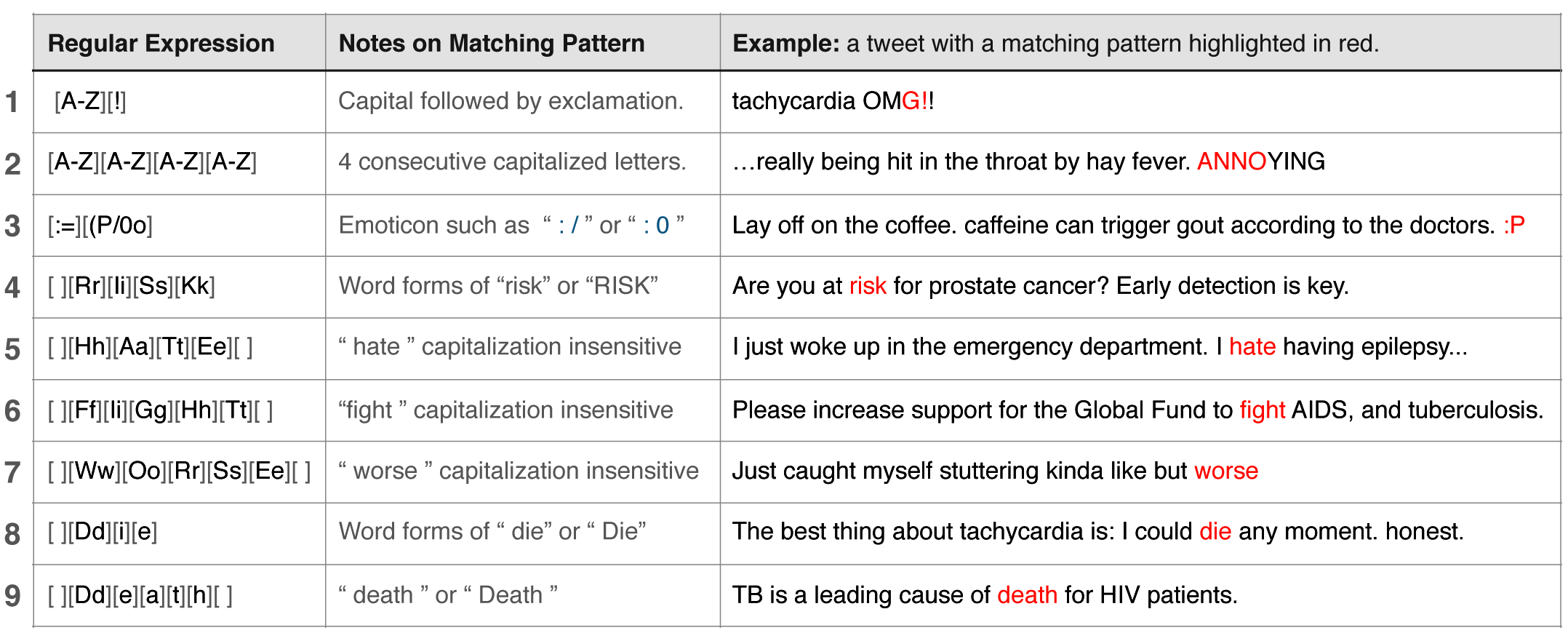}	
\caption{Features}	
\end{center}
\vspace{-8.5mm}
\end{table}

For the second approach, we relied on large scale screening to select single words (unigrams, hereafter). As a convenient source, we took a sentiment dictionary containing 2477 unigrams used for sentiment analysis (SA),\cite{nielsen_new_2011} then found the frequency of these unigrams in our social media corpus. Selecting the most frequent 50 unigrams, we further refined this set to include 16 unigrams that best separated the classes.

While we screened more than 60 variables for relevance, we selected only 24 features for training, which included 16 unigrams, three emoticon features, and five formality patterns. Using backward selection, we eliminated 15 features that had a minimal contribution to the classification accuracy on the validation set, leaving nine features in the final algorithm.

\begin{wraptable}{r}{0.65\textwidth}
\vspace{-4mm}
\captionsetup{font=scriptsize}
\includegraphics[width = .65\textwidth]{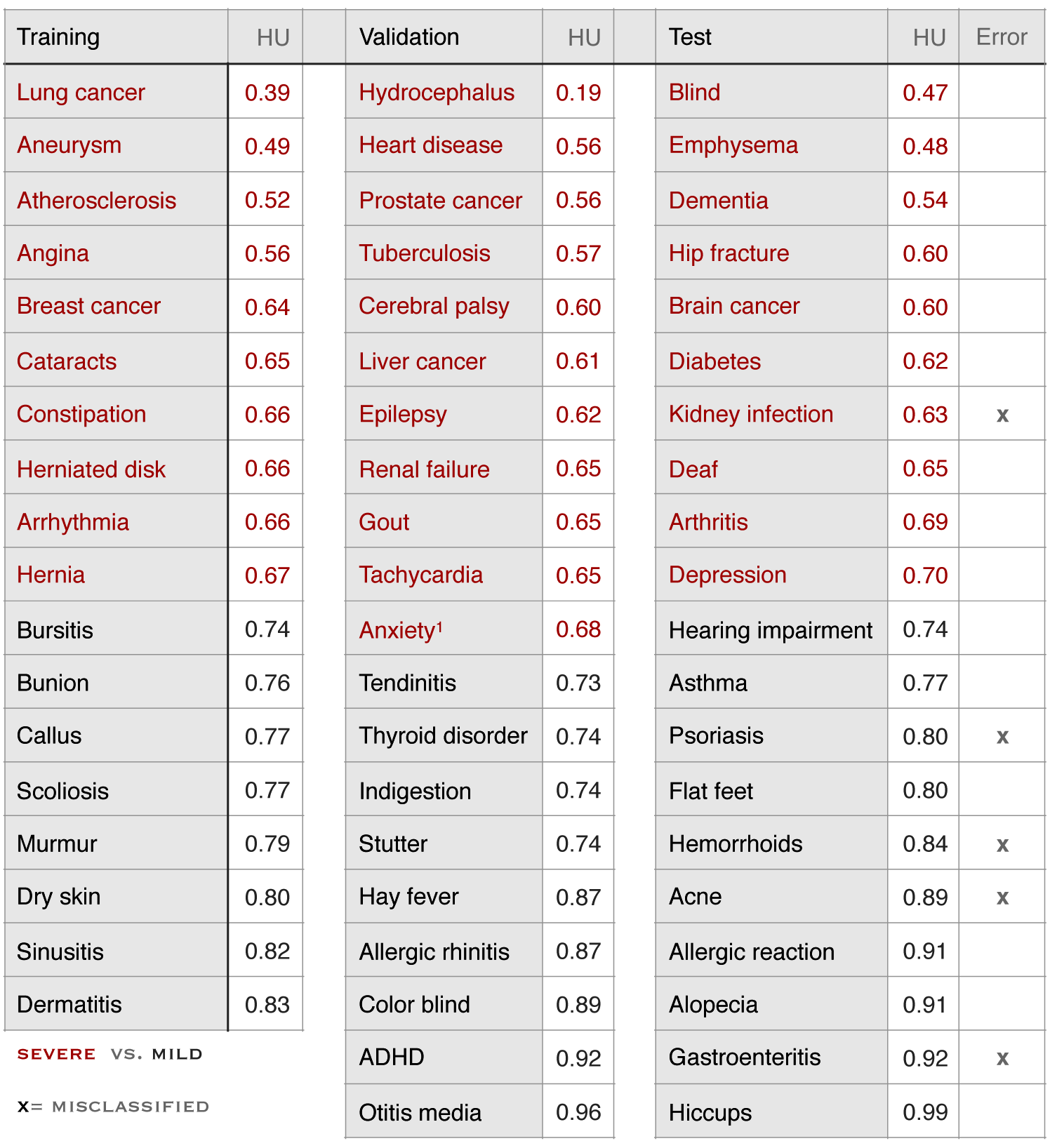}	
\caption{Health Conditions}	
\vspace{-6mm}
\end{wraptable}	
For our learning algorithms, we compared the following classifiers implemented in the Python library, Scikit-learn: logistic regression, a decision tree, a linear support vector machine, and a Random Forest. These algorithms represent widely known, general-purpose algorithms in machine learning.

The decision tree classifier (with default settings) performed the best of the four algorithms: its accuracy is 72\% on the test set. The final algorithm includes nine features, including emoticons, unigrams, and patterns indicating formality of sentence construction. The algorithm tends to misclassify mild diseases as severe. Amid these misclassified examples, there are no obvious trends in terms of data characteristics, such as number of tweets in the test set, nor are there obvious clinical features linking these diseases, such as age of onset, duration, or gender distribution.

Thus we have shown modest accuracy despite several challenges: our social media content is wide-ranging, and users are only a subset of the population. We did not extensively filter tweets to narrow context, such as for self-experience versus unrelated content. Also, the volume of tweets on each health condition is quite variable, ranging from about 100 tweets to 200,000. Since we had to manually extract suitable HUs from the literature, we have a limited supply of examples for training and validation.

\begin{wraptable}{l}{0.35\textwidth}
\vspace{-3mm}
\captionsetup{font=scriptsize}
\includegraphics[width = .35\textwidth]{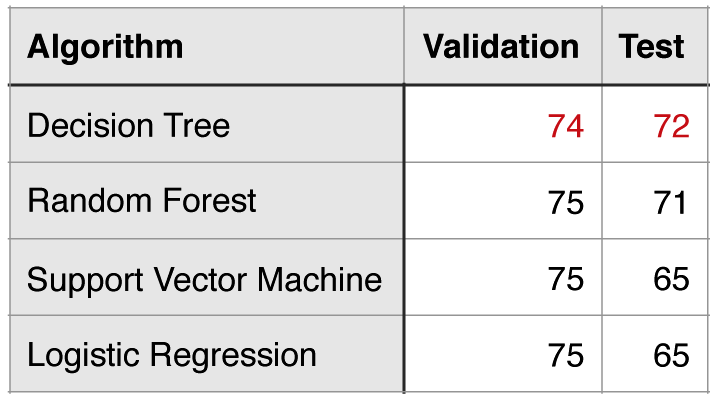}	
\caption{Accuracy}	
\vspace{-3mm}
\end{wraptable}

Limitations of data mining include the following: as noted in similar studies of online patient generated data, periodic crowd-level events cause unpredictability.\cite{ginsberg_detecting_2009} Using a diverse set of predictive features, and filtering for specific contexts could mitigate this to some degree. Also, attention will have to be paid to who is posting messages to social media and how they advocate or otherwise act as a proxy for patients with a particular disease.  Sometimes HUs are obtained within highly controlled situations, such as clinical drug trials; clearly data mining cannot replace face-to-face elicitation in these circumstances.

Data mining addresses the many gaps in knowledge that exist (especially for the developing world) because of the expense of the traditional methodology. Our work represents a movement towards an information-age scaling of a clinical, research, and quality improvement task: gauging patient complaints related to health conditions and outcomes. In the business world, data mining of consumer sentiment is already widespread. Epidemiologists have already developed systems for tracking case counts using search engine data,\cite{ginsberg_detecting_2009} Twitter,\cite{lamb_separating_2013} or Wikipedia pageviews.\cite{generous_global_2014} Much of this work has focused on case counts but has lacked an assessment of case severity, and therefore misses an aspect of disease burden.

In summary, we have used 11 million health-related tweets and a learning algorithm to classify health conditions as severe or mild, a necessary step in the final goal of placing conditions on a continuum of severity. The power of data mining is that it can be applied to any disease with measurements occurring at any time interval. Unlike a survey, which has been the mainstay for more than 70 years, the marginal cost of analyzing an additional health condition is virtually zero. In particular, this has implications for the developing world, where text messaging and social media have grown increasingly prevalent and yet the assessment of disease burden remains challenging.

\section{Methods}
\noindent\textbf{Literature Selection:} We used Pubmed and Google Scholar to identify established studies that defined health utilities for multiple, specific health conditions. Most of our HUs were obtained from a single study,\cite{gold_toward_1998} sponsored by the National Center for Health Statistics (NCHS). It met our criteria since it included 130 conditions and (as an indicator of acceptance) had 268 citations as of April 23, 2016. This agency, charged with monitoring the health of the US population for the Department of Health and Human Services, conducted the research to validate a novel HU measurement instrument (called HALex). The authors of the NCHS study conclude that HALex has good face validity and high correlations to other canonical studies. A secondary goal was to generate a catalog of HUs useful to studies that must rely on HU information from secondary data sources due to the prohibitive expense of acquiring primary data. We bolstered our HU dataset with other studies cataloging HUs,\cite{fryback_beaver_1993} including two containing pediatric health conditions\cite{carroll_improving_2009, petrou_estimating_2009} in order to diversify the conditions studied. A few conditions were included that were not part of studies assessing multiple health conditions simultaneously.\cite{bilgic_psychiatric_2014, fluchel_self_2008, roux_burden_2012}  To mitigate errors related to methodological variations between studies, we preferred that these single-study conditions had clear class membership (very mild or very severe).  \\

\noindent \textbf{Selection of Concepts:}  Although the NCHS study reports HUs on many specific illnesses, it still contains HUs associated with unhelpful, broader diagnostic bins: examples include `other extremity paralysis', `orthopedic impairment-other', `absent bone/joint', and `hand or finger impairment'. We used common experience and the medical expertise of one of the authors, CLT, a medical doctor, to filter non-specific terms. In about 30 cases, we could not use a selected concept since we had an insufficient number of tweets (arbitrarily, fewer than 100). \\

\noindent\textbf{Selection of Tweets:} We filtered our dataset of 11 million health-related tweets for content related to our research. First, we selected all the tweets containing our concept of interest, matching tweets with a capitalization-insensitive algorithm. A concept is comprised of multiple descriptions (synonyms). For each HU study, we found a summary table listing every health condition measured and its corresponding HU. We extracted the exact health description mentioned in the summary table, and then retrieved all of its descriptions and pertinent word forms as defined by a consumer health vocabulary mentioned above. If a tweet matched more than one concept, it was selected for each concept.\\

\noindent\textbf{Privacy:} The health-related tweets obtained for this study contained only the content of the tweet, but we eliminated all other metadata. Because the tweets were intentionally public, this study was exempt from the IRB process. The tweets in table 1 were also public, and have been further modified by us to obscure attribution to a specific user.\\

\noindent\textbf{Algorithm Development:}  For our learning algorithms, we compared the classifiers as discussed in the body of this article and implemented in the Python library, Scikit-learn. We randomized the examples to training, validation and test sets using stratified randomization, so that each set contained a similar range of health utilities. As discussed, we sequentially explored the decision tree hyperparameters by simply assigning increasingly extreme values while monitoring the effect on the validation error. As discussed above, we selected candidate predictive features via two approaches: manually developing features that capture differences in formality and in the second approach, differentiating frequencies of unigrams from the affective lexicon. Nuances related to the capitalization of these unigrams change their predictive value. \\ 		
	
\section{References}

\bibliography{bibtex_HU}


\section{End Notes}
\noindent\textbf{Supplementary Information} 
Table comprised of all health conditions used in the study and for each: number of tweets, HU value from literature, source of HU value, partition (training, validation, or test).

\noindent\textbf{Acknowledgements} 
The authors are grateful to Michael J. Paul and Mark Dredze for sharing Tweets, Reid Priedhorsky for sharing feedback, and Rebecca Anthony for her editing, critiques, and encouragement.

\noindent\textbf{Author Contributions:} 
C.L.T. conceptualized the study. J.I. provided feedback on the study design.  C.L.T. and C.Y. performed the analyses. C.L.T. prepared the manuscript, with extensive editing by J.I. and C.Y.

\noindent\textbf{Author Information:} 
The authors are affiliated with Michigan State University, (East Lansing, MI 48824, USA), from the following departments: Chris Thompson, MD, MHI is from pediatrics, Clint Young, PhD is from the department of Physics, Astronomy and the National Superconducting Cyclotron Laboratory. Josh Introne, PhD is from Media and Information. The authors declare no competing financial interests.  The corresponding author is Christopher L. Thompson.

\end{document}